\documentclass[lettersize,journal]{IEEEtran}
\usepackage{amsmath}
\usepackage{amsfonts,bm}
\usepackage{amsthm}
\usepackage{algorithm,algorithmicx,algpseudocode}
\usepackage{array}
\usepackage{textcomp}
\usepackage{stfloats}
\usepackage{url}
\usepackage{verbatim}
\usepackage{graphicx}
\usepackage{cite}
\usepackage{multirow}
\usepackage{booktabs}
\usepackage{hyperref}
\usepackage{stmaryrd}
\usepackage{amssymb}
\usepackage{pifont}
\usepackage{subcaption}
\usepackage{boldline}
\usepackage{cite}
\usepackage{enumitem}
\usepackage{calc}
\usepackage{multibib}
\usepackage{nomencl}
\usepackage{etoolbox}
\usepackage{mathtools}
\usepackage{cleveref}

\usepackage[textsize=footnotesize,textwidth=17mm]{todonotes}

\makenomenclature
\renewcommand{\nompreamble}{
The next list describes several symbols that will be later used.  
Sets and indices are in calligraphic or blackboard bold font. 
Bold symbols represent vectors or matrices. 
}
\renewcommand\nomgroup[1]{%
  \item[\bfseries
  \ifstrequal{#1}{V}{Variables}{%
  \ifstrequal{#1}{P}{Parameters}{%
  \ifstrequal{#1}{M}{Machine learning related}{%
  \ifstrequal{#1}{S}{Set and indices}{}}}}%
]}

\newcommand{\tb}[1]{\textcolor{black}{#1}}

\newcommand{\mmin}{\operatornamewithlimits{\text{\textbf{min}}}}

\newcommand{\cG}{\mathcal{G}} 
\newcommand{\cE}{\mathcal{E}} 
\newcommand{\cN}{\mathcal{N}} 
\newcommand{\cL}{\mathcal{L}} 
\newcommand{\slack}{\bm{\eta}} 
\newcommand{\slackapp}{\tilde{\bm{\eta}}}
\newcommand{\load}{\mathbf{d}} 
\newcommand{\g}{\mathbf{g}} 
\newcommand{\f}{\mathbf{f}} 
\newcommand{\gk}{\mathbf{g}_k} 
\newcommand{\gkprov}{\mathbf{g}'_k} 
\newcommand{\rhok}{\bm{\rho}_k} 
\newcommand{\nk}{n_k} 

\newcommand{\Kg}{\mathcal{K}_g} 
\newcommand{\Ke}{\mathcal{K}_e}
\newcommand{\Kgsub}{\mathbb{K}_g} 
\newcommand{\Ug}{\mathbb{U}_g}
\newcommand{\Ue}{\mathbb{U}_e}

\newcommand{\PTDF}{\mathbf{K}} 
\newcommand{\LODF}{\mathbf{L}} 
\newcommand{\B}{\mathbf{B}} 
\newcommand{\flb}{\underline{\mathbf{f}}} 
\newcommand{\fub}{\overline{\mathbf{f}}} 
\newcommand{\glb}{\underline{\mathbf{g}}} 
\newcommand{\gub}{\overline{\mathbf{g}}} 
\newcommand{\slackpen}{M_\eta} 
\newcommand{\objc}{\mathbf{c}} 

\newcommand{\gammavec}{\bm{\gamma}}
\newcommand{\gcapa}{\hat{\mathbf{g}}}

\newcommand{\x}{\mathbf{x}}
\newcommand{\pnet}{P_{\theta}}
\newcommand{\dnet}{D_{\phi}}
\newcommand{\y}{\mathbf{y}}
\newcommand{\yapp}{\tilde{\mathbf{g}}}
\newcommand{\yappp}{\check{\mathbf{g}}}
\newcommand{\gkapp}{\tilde{\mathbf{g}}_k} 
\newcommand{\obj}{f_{\x}}

\newcommand{\eq}{\mathbf{h}_{\x}}
\newcommand{\deq}{\bm{\lambda}}

\newcommand{\viol}[1]{\nu\left(#1\right)}

\newcommand{\rhomax}{\rho_{\text{max}}}
\newcommand{\dataset}{\mathcal{D}}

\newcommand{\norm}[1]{\left\lVert#1\right\rVert} 
\newcommand{\linfnorm}[1]{\left\lVert#1\right\rVert_\infty} 

\newfloat{model}{thp}{lop}
\floatname{model}{Model}

\begin{document}

\title{Self-Supervised Learning for Large-Scale Preventive Security Constrained DC Optimal Power Flow}

\author{Seonho~Park and Pascal~Van~Hentenryck
    \thanks{
        The authors are affiliated with the School of Industrial and Systems Engineering, Georgia Institute of Technology, Atlanta, GA 30332, USA, 
        E-mail: seonho.park@gatech.edu, pascal.vanhentenryck@isye.gatech.edu
    }
}

\maketitle

\begin{abstract}
Security-Constrained Optimal Power Flow (SCOPF) plays a crucial role in power grid stability but becomes increasingly complex as systems grow.  
This paper introduces Primal-Dual Learning (PDL) for SCOPF (PDL-SCOPF), a self-supervised end-to-end primal-dual learning framework for producing near-optimal solutions to large-scale SCOPF problems in milliseconds.  
Indeed, PDL-SCOPF remedies the limitations of supervised counterparts that rely on training instances with their optimal solutions, which becomes impractical for large-scale SCOPF problems. 
PDL-SCOPF mimics an Augmented Lagrangian Method (ALM) for training primal and dual networks that learn the primal solutions and the Lagrangian multipliers, respectively, to the unconstrained optimizations. 
In addition, PDL-SCOPF incorporates a repair layer to ensure the feasibility of the power balance in the nominal case, and a binary search layer to compute, using the Automatic Primary Response (APR), the generator dispatches in the contingencies. 
The resulting differentiable program can then be trained end-to-end using the objective function of the SCOPF and the power balance constraints of the contingencies. 
Experimental results demonstrate that the PDL-SCOPF delivers accurate feasible solutions with minimal optimality gaps. 
The framework underlying PDL-SCOPF aims at bridging the gap between traditional optimization methods and machine learning, highlighting the potential of self-supervised end-to-end primal-dual learning for large-scale optimization tasks.
\end{abstract}

\begin{IEEEkeywords}
Primal-dual learning, Security-constrained optimal power flow, Column and constraint generation, Deep learning, Differentiable programming, End-to-end learning
\end{IEEEkeywords}
\nomenclature[V]{$\g$}{Power generations (dispatches, injections)}
\nomenclature[V]{$\slack$}{Slack variable for power flow limit constraints}
\nomenclature[V]{$\f$}{Base case power flow vector}
\nomenclature[V]{$\f_k$}{Power flow vector under the generator contingency $k$}
\nomenclature[V]{$\gk$}{Power generations under the generator contingency $k$}
\nomenclature[V]{$\gkprov$}{Provisional variables for $\gk$}
\nomenclature[V]{$\rhok$}{Binary variable indicating whether $\gk$ reaches the upper limit under the generator contingency $k$}
\nomenclature[V]{$\nk$}{Global signal variable under the generator contingency $k$}

\nomenclature[S]{$\cG,\cE,\cN,\cL$}{Indices of the generators, transmission lines, buses, load units}
\nomenclature[S]{$\Kg,\Ke$}{Generator or transmission line contingency indices}
\nomenclature[S]{$\Kgsub$}{Subset of $\Kg$ considered in the current CCGA iteration}
\nomenclature[S]{$\Ug,\Ue$}{Sets of pairs of $(k,l)$ where the power flow limit constraint at line $l$ is violated under the generator or line contingency $k$}

\nomenclature[M]{$\slackapp$}{Slack variable estimate}
\nomenclature[M]{$\x$}{Input parameters}
\nomenclature[M]{$\pnet$}{Primal network}
\nomenclature[M]{$\dnet$}{Dual network}
\nomenclature[M]{$\yapp$}{Base case dispatch estimate}
\nomenclature[M]{$\yappp$}{Base case dispatch estimate before applying the power balance repair layer}
\nomenclature[M]{$\gkapp$}{Dispatch estimate under the generator contingency $k$}
\nomenclature[M]{$\deq$}{Dual variable estimate associated with the power balance constraint under the generator contingencies}

\nomenclature[P]{$\load$}{Load demands}
\nomenclature[P]{$\PTDF$}{Power transfer distribution factor matrix}
\nomenclature[P]{$\LODF$}{Line outage distribution factor matrix}
\nomenclature[P]{$\B$}{Generator--bus incidence matrix}
\nomenclature[P]{$\flb,\fub$}{Lower and upper bound of power flow limit}
\nomenclature[P]{$\glb,\gub$}{Lower and upper bound of generator dispatch}
\nomenclature[P]{$\slackpen$}{Penalty coefficient}
\nomenclature[P]{$\objc$}{Cost coefficients}
\nomenclature[P]{$\gammavec$}{Primary response parameter}
\nomenclature[P]{$\gcapa$}{Generator capacities}
\nomenclature[P]{$\delta$}{Relaxation parameter for the binary search layer}
\nomenclature[P]{$\beta$}{Threshold parameter for choosing the cuts to be added in the CCGA}

\printnomenclature

\section{Introduction}
\label{sec:intro}
Power system operations must maintain the power balance between load
and generation at all times. To achieve this equilibrium requires
solving mathematical optimization problems in real-time and
day-ahead electricity markets. These optimizations have become
increasingly challenging in recent years due to the integration of
renewable energy sources which are both more numerous and volatile
compared to traditional synchronous generators. As a consequence,
managing the reliability and the risk in the system gets increasingly
important.

The Security-Constrained Optimal Power Flow (SCOPF) problem is a
traditional model to ensure the stability of power systems operations
under contingencies. The SCOPF guarantees that a feasible generator
dispatch exists even under the failure of a single generator or
transmission line, while minimizing the cost of the nominal
dispatch. In actual operations, however, Independent System Operators
(ISOs) do not solve the SCOPF; rather they maintain reserves that are
computed outside the markets; the reserve requirements may be
sub-optimal or insufficient in cases with large shares of
renewables. One reason for using reserves is the high computational
cost of the SCOPF compared to its equivalent Economic Dispatch (ED)
with reserves, even when using a DC approximation of the power flow
equations. Indeed, the DC-SCOPF is a Mixed-Integer Linear Program (MILP) 
whose number of binary variables increases quadratically with the number of
contingencies for preventive SCOPF with Automatic Primary generator
Response (APR) \cite{dvorkin2016optimizing}. In contrast, the ED with
the reserve constraints is a linear program that can be readily solved.

One possible approach to address the computational challenges of the
SCOPF is the use of Machine Learning (ML) which has attracted
significant attention in the power systems community in recent
years. However, even for machine learning, the SCOPF raises fundamental
challenges since generating sufficient training data is typically
impractical for industrial-size power grids. Moreover, the ML models
may have to predict a quadratic number of variables to ensure the
power balance constraints in the contingencies.

This paper takes a different avenue. It considers the ED 
formulation used by the ISOs in the United States where the
reserve requirements have been replaced by $N\!-\!1$ contingencies to
produce a preventive APR-based SCOPF. The paper then proposes an {\em
  end-to-end self-supervised primal-dual Learning framework}, called
PDL-SCOPF, to address the computational challenges raised by the SCOPF
formulation. By virtue of being self-supervised, PDL-SCOPF does not
need the optimal solutions of thousands of training instances: it just
relies on the availability of an empirical distribution of the input configuration
that captures future conditions of interest. 
Thanks to being trained end-to-end,
PDL-SCOPF only needs to predict the baseline dispatch: differentiable
layers are then used to restore the feasibility of the power balance
for the base case and to predict the contingency dispatches using the
APR. PDL-SCOPF employs a Primal-Dual Learning (PDL) framework whose
training mimics the Augmented Lagrangian Method (ALM) over a set of
instances: it uses a primal network to approximate unconstrained
optimization problems similar to those of the ALM and a dual network
to approximate the Lagrangian multipliers used in these optimizations.

The main contribution of the paper, PDL-SCOPF, can be summarized as follows:
\begin{itemize}
\item PDL-SCOPF is a self-supervised method that produces near-optimal
  feasible solutions to preventive DC-SCOPF. Thus, it does not require 
  a training dataset that contains optimal SCOPF solutions for a set of instances.
\item PDL-SCOPF is a Primal-Dual Learning framework that mimics Augmented
  Lagrangian Method by predicting both primal optimal solutions (using a primal network) and their associated Lagrangian multiplier (using a dual network).
\item PDL-SCOPF uses differential layers to restore the feasibility of the
  power balance \cite{chen2022learning} and to adapt the binary search
  from the Column and Constraint Generation Algorithm in
  \cite{velloso2021exact} to compute the contingency dispatches and
  their violations. The predictive model and these differentiable layers can then be trained
  end-to-end in the framework to produce near-optimal feasible solutions.
\item The scalability and accuracy of PDL-SCOPF have been validated on industry-size test cases
  with thousands of buses, for which it estimates solutions in milliseconds. 
\end{itemize}

\noindent
The rest of the paper is structured as follows. Section
\ref{section:related} describes the related work in SCOPF and machine
learning. Section~\ref{sec:prob} revisits the formulation of SCOPF and
briefly describes a Column and Constraint Generation Algorithm to
solve it exactly. Section~\ref{sec:model} revisits the PDL framework
and Section~\ref{section:PDL-SCOPF} introduces PDL-SCOPF.
Section~\ref{sec:exp} presents the experimental results and compares
PDL-SCOPF with a number of baselines.  Section~\ref{sec:conclusion}
summarizes the paper and presents directions for future work.

\section{Related Work}
\label{section:related}

\paragraph{Security-Constrained OPF}
This paper focuses on the SCOPF problem with $N\!-\!1$ generator and line
contingencies. A comprehensive review of SCOPF can be found in
\cite{capitanescu2011state}. The SCOPF aims at determining the
pre-contingency generator dispatch that minimizes operational costs
while ensuring feasibility even in the event of contingencies.  A $N\!-\!1$
contingency refers to the outage of any single component (generator
or line). The SCOPF can be classified into two distinct settings: 1)
the corrective case, discussed in \cite{wang2016solving}, which
assumes that re-scheduling is possible, and 2) the preventive case,
studied for instance in
\cite{li2008decomposed,ardakani2013identification,dvorkin2016optimizing,velloso2021exact},
where re-dispatch is not an option. This paper considers a preventive
SCOPF that models an APR for generator
contingencies. Such SCOPF problems are computationally challenging due
to the binary variables introduced to model the APR response. Indeed, the
number of binary variables increases quadratically in the number of
generator contingencies. To alleviate this computational burden,
various decomposition techniques have been proposed: they include
Benders decomposition (e.g., \cite{bertsimas2012adaptive}, and the
Column and Constraint Generation Algorithm (CCGA) proposed in
\cite{zeng2013solving,velloso2021exact}. The PDL-SCOPF method
proposed in this paper uses some insights from the CCGA. 
Extensive reviews on AC-SCOPF are available in
\cite{platbrood2013generic,capitanescu2016critical}.  Recent efforts
to tackle AC-SCOPF, as part of the Grid Optimization Competition
initiated by ARPA-E, are summarized in \cite{aravena2022recent}. Both
industry and academia have also explored approximations and
relaxations to the AC-SCOPF, as discussed in
\cite{coffrin2014linear,coffrin2015qc}.  Currently, industry (e.g.,
ISO) predominantly employs the DC
formulation for SCOPF (DC-SCOPF) \cite{marano2012exploiting}.
Consequently, this paper focuses on the DC-SCOPF. More precisely, the
formulation presented in this paper can be viewed as the economic
dispatch of the ISOs where the reserve constraints have been replaced
by $N\!-\!1$ contingencies under the APR modeling.
For a broader perspective, see the recent comprehensive review on
large grid optimization in \cite{pandey2023large}.

\paragraph{Machine Learning for Optimization and Power Systems}
Various end-to-end Machine Learning (ML) approaches have recently
emerged for optimization applications. They aim at directly
estimating optimal solutions given a distribution of the input
parameters. In other words, these ML approaches are used as regressions 
to approximate the mapping from input parameters to optimal
solutions.  A comprehensive overview of end-to-end optimization
learning methods can be found in \cite{kotary2021end}.  A ML framework to
accelerate general mixed integer programming (MIP) solvers was
proposed in \cite{nair2020solving}; it also offers a way to bound the
optimality gap between the inference and the optimal solution.
End-to-end optimization learning techniques have also founded
applications in various power system optimization tasks, such as
economic dispatch \cite{chen2022learning}, DCOPF
\cite{pan2020deepopf}, ACOPF \cite{fioretto2020predicting}, and unit
commitment \cite{xavier2021learning, park2022confidence}.

\paragraph{Supervised Learning for SCOPF}
Several supervised learning approaches have been applied to SCOPF.
{\sc DeepOPF}~\cite{pan2020deepopf} approximates the optimal dispatch
of the SCOPF problem, taking into account line contingencies.  Velloso
and Van Hentenryck~\cite{velloso2021combining} combine a deep neural
network-based mapping with the CCGA method to approximate the optimal
solution of the SCOPF problem, with a focus on generator
contingencies.  These approaches employed supervised learning, which
requires gathering pairs of input parameters and their corresponding
optimal solutions for training. 
As demonstrated in Section~\ref{sec:exp}, solving a
single SCOPF instance for large-scale SCOPF is quite time-consuming,
which makes the collection of such training data impractical.  This
paper remedies this limitation by using a self-supervised learning
framework that does not necessitate the ground truth solutions for 
training, which, in turn, is especially effective for learning 
large-scale industry-sized SCOPF problems having both generator 
and line contingencies.

\paragraph{Self-Supervised Learning for OPF}
Self-supervised learning leverages the original optimization
formulation, including the objective function, to approximate optimal
solutions without relying on the ground truth data.  DC3
\cite{donti2021dc3} is such a self-supervised method that employs an
implicit layer to driving the learning towards feasible solutions.
E2ELR \cite{chen2022learning} introduces repair layers for satisfying 
power balance and reserve constraints that are trained end-to-end 
to produce feasible and near-optimal solutions to
ED problems. PDL
\cite{park2023self} integrates primal and dual networks to approximate
both the primal and dual solutions for constrained optimization
problems.  In this work, PDL-SCOPF leverages the E2ELR repair layer for power
balance, PDL, and the binary search layer adapted from CCGA to produce
near-optimal feasible solutions to industry-size SCOPF problems.

\paragraph{Scalable Learning for OPF} 
Significant work has been carried out for scaling ML
approach to large-scale OPF. 
They include Compact Learning \cite{park2023compact},
which learns a lower-rank representation of the optimal solution
for ACOPF, and Spatial Decomposition \cite{chatzos2021spatial},
which proposes a two-step learning process by learning the flows
between regions in a spatial decomposition before learning the
optimal power flow in each region.

\section{Problem Formulation and Objective}
\label{sec:prob}
This section reviews the SCOPF problem considered in this paper. The
formulation can be seen as the traditional economic dispatch of the
ISOs in the United States, where the reserve
constraints have been replaced by an explicit modeling of $N\!-\!1$
contingencies. In the formulation, the thermal limits are thus
considered as soft constraints but their violations are heavily
penalized in the objective function. To handle contingencies, the
formulation captures the Automatic Primary Response (APR)
\cite{dvorkin2016optimizing,aravena2022recent} used in
\cite{velloso2021exact,velloso2021combining}.

\subsection{The SCOPF Problem}
\label{ssec:SCOPF}

\begin{model}[!t]
{\scriptsize
\caption{The Extensive SCOPF Formulation}
\label{model:scopf_ext}
\begin{flalign}
    \:\:\:\:\:\:\:\:
    \:\:\:\:\:
    \mmin_{
    \mathclap{
    \substack{ \setlength{\jot}{-0.8\baselineskip}\everymath{\scriptstyle}
    \begin{array}{c}
    \g,\\ 
    \!\!\left[\gk\!,\rhok\!,\nk\!\right]_{k\in\Kg}\!, \\
    \!\!\left[\slack_k\right]_{\!k\in\{\!0\!\}\cup\Kg\cup\Ke}
    \end{array}
  }
  }
  }
 \:\:\:\:& \objc^{\top}\g \!+\! \slackpen\!\!\left(\sum_{k\in\{\!0\!\}\cup\Kg\cup\Ke}\!\!\norm{\slack_k}_{\!1} \!\!\right)\label{eq:scopf_ext_obj}\\   
\text{\textbf{s. t.}:}\:\:\:\:\:& \mathbf{1}^\top\g = \mathbf{1}^\top\load \label{eq:scopf_ext_cnst_pb} \\
 & \flb\!-\!\slack_0 \leq \f\!=\!\PTDF (\load\!-\!\B\g) \leq \fub\!+\!\slack_0  \label{eq:scopf_ext_cnst_pf} \\
 & \glb \leq \g \leq \gub \label{eq:scopf_ext_cnst_gbound} \\
 & \mathbf{1}^\top\gk = \mathbf{1}^\top\load &\!\!\!\!&\!\!\!\! \forall k\!\in\!\Kg \label{eq:scopf_ext_cnst_pb_kg} \\
 & \flb\!-\!\slack_k \!\leq\! \f_k\!=\!\PTDF (\load\!-\!\B\gk) \!\leq\! \fub\!+\!\slack_k&\!\!\!\!&\!\!\!\!\forall k\!\in\!\Kg \label{eq:scopf_ext_cnst_pf_kg} \\
 & \underline{g}_i \leq g_{k\!,i} \leq \overline{g}_i                  &\!\!\!\!&\!\!\!\! \forall i\!\in\!\cG,\forall k\!\in\!\Kg,i\!\neq\! k \label{eq:scopf_ext_cnst_gbound_kg} \\
 & g_{k\!,k} = 0 &\!\!\!\!&\!\!\!\! \forall k\!\in\!\Kg \label{eq:scopf_ext_cnst_gbound_kg2} \\
 & |g_{k\!,i}\!-\!g_i\!-\!n_k\gamma_i\hat{g}_i|\!\leq\!\hat{g}_i\rho_{k\!,i} &\!\!\!\!&\!\!\!\! \forall i\!\in\!\cG,\forall k\!\in\!\Kg,i\!\neq\! k \label{eq:scopf_ext_cnst_apr1} \\
 & g_i\!+\!n_k\gamma_i\hat{g}_i\!\geq\!\hat{g}_i\rho_{k\!,i}\!+\!\underline{g}_i &\!\!\!\!&\!\!\!\! \forall i\!\in\!\cG,\forall k\!\in\!\Kg,i\!\neq\! k \label{eq:scopf_ext_cnst_apr2} \\
 & g_{k\!,i} \!\geq\!\hat{g}_i\rho_{k\!,i}\!+\!\underline{g}_i &\!\!\!\!&\!\!\!\! \forall i\!\in\!\cG,\forall k\!\in\!\Kg,i\!\neq\! k \label{eq:scopf_ext_cnst_apr3} \\
 & \flb\!-\!\slack_k \leq \f \!+\! f_k \LODF_k  \leq \fub\!+\!\slack_k &\!\!\!\!&\!\!\!\!\forall k\!\in\!\Ke \label{eq:scopf_ext_cnst_pf_ke} \\
 & \slack_k \geq 0 &\!\!\!\!&\!\!\!\!\forall k\!\in \{\!0\!\}\!\cup\!\Kg\!\cup\!\Ke \label{eq:scopf_ext_cnst_slack_bound} \\ 
 & \nk \in \left[0,1\right] &\!\!\!\!&\!\!\!\!\forall k\!\in\!\Kg \label{eq:scopf_ext_cnst_n_kg_bound} \\
 & \rho_{k\!,i} \in \left\{0,1\right\} &\!\!\!\!&\!\!\!\!\forall i\!\in\!\cG,\forall k\!\in\!\Kg,i\!\neq\! k \label{eq:scopf_ext_cnst_rho_kg_bound}
\end{flalign}
}
\end{model}

Model~\ref{model:scopf_ext} presents the extensive SCOPF problems
with $N-1$ generator and line contingencies.  Its primary objective is
to determine the optimal active generation points for the base case
while ensuring the feasibility of both generator and line
contingencies.  Objective~\eqref{eq:scopf_ext_obj} sums the linear
cost of the base case dispatch $\g$ and the penalties for violating
the thermal limits in the base case and in the contingencies, with
$\slackpen$ being a large penalty coefficient.

\paragraph{The Base Case}

The base case includes Constraints~\eqref{eq:scopf_ext_cnst_pb},
\eqref{eq:scopf_ext_cnst_pf}, and \eqref{eq:scopf_ext_cnst_gbound} that
should be satisfied.  Constraint~\eqref{eq:scopf_ext_cnst_pb} imposes
the power balance between the generation and the load demands in the
nominal case. In this constraint, $\mathbf{1}$ represents a vector whose
elements are all one, and $\load\in\mathbb{R}^{\lvert\cN\rvert}$ is
the load demand with the bus-wise representation of the load units.
Constraint~\eqref{eq:scopf_ext_cnst_pf} ensures that the power flow
$\f$ for each transmission line falls within the predefined upper and
lower limits, $\flb$ and $\fub$.  Any violation is penalized in the
objective value.  The power flow for the base case 
$\f\!=\!\PTDF (\load\!-\!\B\g)$ is
computed using a Power Transfer Distribution Factor (PTDF) matrix
$\PTDF\in\mathbb{R}^{|\cE|\times|\cN|}$ and a generator-bus incidence
matrix $\B$. Constraint~\eqref{eq:scopf_ext_cnst_gbound} imposes the
generation limits.

\paragraph{Generator Contingency}

Each generator contingency imposes
Constraints~\eqref{eq:scopf_ext_cnst_pb_kg},
\eqref{eq:scopf_ext_cnst_pf_kg}, and
\eqref{eq:scopf_ext_cnst_gbound_kg} to enforce the power balance, the
thermal limits, and the generation bounds under the generator contingency.  
The only difference from the base case is Constraint~\eqref{eq:scopf_ext_cnst_gbound_kg2} 
that specifies that generator $k$ should remain inactive under its contingency.

\paragraph{Automatic Primary Response for Generator Contingency}
To address generator contingencies,
Constraints~\eqref{eq:scopf_ext_cnst_apr1}--\eqref{eq:scopf_ext_cnst_apr3}
implements an APR
\cite{dvorkin2016optimizing,aravena2022recent}. The formulation in this paper is
based on the APR model used in
\cite{velloso2021exact,velloso2021combining} where, for each generator
contingency $k$, a system-wide signal $n_k \in[0,1]$ represents the
level of system response required to resolve the power imbalance.
Furthermore, the APR model assumes that the change in synchronized
generation dispatch under a contingency is proportionate to the droop
slope, which is determined by the product of generator capacity
$\hat{g}$ and the predefined parameter $\gamma$ as in
\cite{velloso2021exact}.  The generator capacities are defined as
$\gcapa\!=\!\gub\!-\!\glb$.  The APR constraints ensure that the
generation dispatch under generator contingency remains within the
generation limits.  The mathematical expression for constraining the
generation dispatch for synchronized generators can be defined as:
\begin{equation}
\label{eq:apr}
    g_{k,i} = \min\{g_i\!+\!n_k\gamma_i\hat{g}_i,\overline{g}_i\},\,\,\forall i\!\in\!\cG, \forall k\!\in\!\Kg,i\!\neq\!k.
\end{equation}
To represent the disjunctive constraint~\eqref{eq:apr}, binary
variables $\rho_{k,i}$ are introduced.  For all $i\in\cG$ and
$k\in\Kg$ such that $k\neq i$, these binary variables enable the
mathematical representation of the above disjunctive
constraint~\eqref{eq:apr}.  The binary variable $\rho$ imposes
$g_k\!=\!\overline{g}$ if $\rho\!=\!1$, and
$g_k\!=\!g+n_k\gamma\hat{g}$ otherwise.  As a result, the extensive
SCOPF problem~\ref{model:scopf_ext}, featuring the APR of generators
under generator contingencies, is a MILP problem.

\paragraph{Line Contingency}
Line contingencies impose
Constraints~\eqref{eq:scopf_ext_cnst_pf_ke}.  These constraints ensure
that, during a line contingency, there is an immediate redistribution
of power flow specified by the Line Outage Distribution Factor (LODF)
\cite{guo2009direct,tejada2017security}.  
Under the line contingency of $k$, $k$-th column vector, denoted as
$\LODF_k$, of the LODF matrix $\LODF\in\mathbb{R}^{|\cE|\times|\Ke|}$,
delineates the redistribution of base case power flow at line $k$, $f_k$,
to the other lines, so as to ensure that there is no power
flow at line $k$, i.e., $f_{k,k}=0,\,\forall k\in\Ke$.

\paragraph{Slack Variables}

The constraints related to thermal limits (\ref{eq:scopf_ext_cnst_pf},
\ref{eq:scopf_ext_cnst_pf_kg}, and \ref{eq:scopf_ext_cnst_pf_ke}) are
treated as \emph{soft constraints}. 
\tb{
Using soft constraints for the thermal limits is in accordance with
the formulations used by ISOs for clearing the electricity markets in
the United States~\cite{ma2009security,BPM_D}.  These soft constraints
capture positive slack variables, as defined
in \eqref{eq:scopf_ext_cnst_slack_bound}, which are heavily penalized
in the objective function.}

\subsection{The Column and Constraint Generation Algorithm (CCGA)}
\label{ssec:ccga}

\begin{algorithm}[!t]
\caption{Column-and-Constraint Generation Algorithm (CCGA) for solving SCOPF}\label{alg:ccga}
{\scriptsize
Initialize: $\Kgsub=\emptyset$, $\Ug=\emptyset$, $\Ue=\emptyset$
\begin{algorithmic}[1]
\For{$j=0,1,\dots$}
    \State Solve Model~\ref{model:CCGA_master} and obtain $\g^{(j)}$
    \State Obtain $\gk^{(j)}$ using the binary search, $\forall k \in \Kg$
    \State Compute the violations $\alpha^g_{k\!,l}$ w.r.t.~\eqref{eq:ccga_master_cnst_pf_kg}, $\forall l\!\in\!\cE,\,\forall k\!\in\!\Kg$
    \State Compute the violations $\alpha^e_{k\!,l}$ w.r.t.~\eqref{eq:ccga_master_cnst_pf_ke}, $\forall l\!\in\!\cE,\,\forall k\!\in\!\Ke$
    \State \textbf{Break} if $\max\{\alpha^g_{k\!,l}\}\leq\epsilon$ and $\max\{\alpha^e_{k\!,l}\}\leq\epsilon$
    \State $\Ug\leftarrow\{(k\!,l)\mid\alpha^g_{k\!,l}>\max\{\alpha^g_{k\!,l}\}/\beta\}$
    \State $\Ue\leftarrow\{(k\!,l)\mid\alpha^e_{k\!,l}>\max\{\alpha^e_{k\!,l}\}/\beta\}$
    \State $\Kgsub\leftarrow\Kgsub\cup\{k \mid \forall k \in \Ug\}$
\EndFor
\end{algorithmic}
}
\end{algorithm}

\begin{model}[!t]
{\scriptsize
\caption{The CCGA Master Problem}
\label{model:CCGA_master}
\begin{flalign}
\:\:\:\:\:\:\:\:\:
\mmin_{
\mathclap{
\substack{ \setlength{\jot}{-0.8\baselineskip}\everymath{\scriptstyle}
\begin{array}{c}
\g,\slack_0,\left[\gkprov\right]_{k\!\in\!\Kg}\!,\\
\left[\rhok\!,\nk\right]_{k\!\in\Kgsub}\!,\\
\!\left[\!\eta_{k\!,l}\!\right]_{\!(k\!,l)\!\in\!\Ug\cup\Ue}
\end{array}
}
}
}
\:\:\:\:& \:\:\:\objc^{\top}\g \!+\! \slackpen\!\!\left(\norm{\slack_0}_{1}+\!\!\!\!\!\!\sum_{(k\!,l)\in\Ug\cup\Ue}\!\!\!\!\!\!\!\!\eta_{k\!,l} \right) \label{eq:ccga_master_obj} \\
\nonumber\text{\textbf{s. t.}:}\:\:\:\:\:\:& \eqref{eq:scopf_ext_cnst_pb},\eqref{eq:scopf_ext_cnst_pf},\eqref{eq:scopf_ext_cnst_gbound}\\
& \slack_0 \geq 0 \label{eq:ccga_master_cnst_slack} \\
& \gkprov-\g \leq \gammavec\gcapa &\!\!\!\!&\!\!\!\! \forall k \!\in\!\Kg \label{eq:ccga_master_cnst_provisional} \\
& \underline{g}_i \leq g'_{k\!,i} \leq \overline{g}_i,                  &\!\!\!\!&\!\!\!\! \forall i\!\in\!\cG,\forall k\!\in\!\Kg,i\!\neq\! k \label{eq:ccga_master_cnst_gbound_kg} \\
& g'_{k\!,k} = 0 &\!\!\!\!&\!\!\!\! \forall k\!\in\!\Kg \label{eq:ccga_master_cnst_gbound_kg2} \\
& \mathbf{1}^\top\gkprov = \mathbf{1}^\top\load &\!\!\!\!&\!\!\!\! \forall k\!\in\!\Kg \label{eq:ccga_master_cnst_pb_kg} \\
& |g'_{k\!,i}\!-\!g_i\!-\!n_k\gamma_i\hat{g}_i|\!\leq\!\hat{g}_i\rho_{k\!,i} &\!\!\!\!&\!\!\!\! \forall i\!\in\!\cG,\forall k\!\in\!\Kgsub,i\!\neq\! k \label{eq:ccga_master_cnst_apr1} \\
 & g_i\!+\!n_k\gamma_i\hat{g}_i\!\geq\!\hat{g}_i\rho_{k\!,i}\!+\!\underline{g}_i &\!\!\!\!&\!\!\!\! \forall i\!\in\!\cG,\forall k\!\in\!\Kgsub,i\!\neq\! k \label{eq:ccga_master_cnst_apr2} \\
 & g'_{k\!,i} \!\geq\!\hat{g}_i\rho_{k\!,i}\!+\!\underline{g}_i &\!\!\!\!&\!\!\!\! \forall i\!\in\!\cG,\forall k\!\in\!\Kgsub,i\!\neq\! k \label{eq:ccga_master_cnst_apr3} \\
& \underline{f}_l\!-\!\eta_{k\!,l} \leq f_{k,l}\leq \overline{f}_l\!+\!\eta_{k\!,l}            &\!\!\!\!&\!\!\!\!\forall (k\!,l)\!\in\!\Ug \label{eq:ccga_master_cnst_pf_kg} \\
& \underline{f}_l\!-\!\eta_{k\!,l} \leq f_l \!+\! (f_k \LODF_{k})_l \leq \overline{f}_l\!+\!\eta_{k\!,l}            &\!\!\!\!&\!\!\!\!\forall (k\!,l)\!\in\!\Ue \label{eq:ccga_master_cnst_pf_ke} \\
& \eta_{k\!,l} \geq 0 & & \!\!\!\!\forall (k\!,l)\in\Ug\!\cup\!\Ue \label{eq:ccga_master_cnst_slack_k}
\end{flalign}
}
\end{model}

Solving the extensive SCOPF model~\ref{model:scopf_ext} directly is extremely
challenging for large networks, as the number of binary variables
grows quadratically with respect to the number of generators.  
The optimization algorithm, CCGA, was proposed in
\cite{velloso2021combining} to address this computational challenge.
This section reviews the CCGA briefly as the PDL-SCOPF framework
proposed in this paper borrows and adapts one of its contributions.

The CCGA is summarized in Algorithm~\eqref{alg:ccga}. It iteratively
solves a master problem that contains a subset of the
contingencies. Additional contingencies with violated constraints are
then identified and added to the master problem after each iteration.
Model~\ref{model:CCGA_master} presents the master problem for the
CCGA, which contains three types of constraints. The first set of
constraints considers those for the base case that are the same as in
the extensive problem~\ref{model:scopf_ext}. The second set
(Constraint~\eqref{eq:ccga_master_cnst_provisional}) is concerned with
the \emph{provisional} generation dispatch $\gkprov$ for each
generator contingency $k$ computed by the master problem.
Constraint~\eqref{eq:ccga_master_cnst_provisional} ensures that
$\gkprov$ always remains within the limits of the APR.  The third set
of constraints considers a subset of generator contingencies $\Kgsub$,
for which binary variables $\rho_{k,i}$ and global signal variable
$n_k$ are defined. It expresses the proportional response on the
provisional dispatch, as well as sets of likely active constraints for
the generator ($\Ug$) and line contingencies ($\Ue$) respectively.

After solving the master problem~\ref{model:CCGA_master}, the CCGA
computes the candidate generation dispatch $\gk^{(j)}$ under generator
contingencies $k$ (where $j$ is the iteration number). 
\tb{ 
Note that the provisional generation dispatch $\gkprov$ is corrected by 
performing the {\em binary search}, introduced in \cite{velloso2021exact},
on $n_k$ for each generator contingency $k$.
This binary search procedure is designed to satisfy the power balance constraints.
Also, note that the purpose of
the provisional generation dispatch $\gkprov$ in the master problem is
to guarantee the existence of the candidate generation dispatches that
satisfy the power balance constraints (as found by the binary search).
Using $\g^{(j)}$ and $\gk^{(j)}$, the CCGA then calculates the
thermal limit violations for all generator and line contingencies.  
If no violations are detected, it means that the current master 
problem has produced an optimal solution, and CCGA stops.  Otherwise, 
$\Ug$ and $\Ue$ are updated by adding constraints with violations 
greater than the threshold $\beta$. CCGA converges in a finite number 
of iterations since it adds the generator contingencies with the 
violated thermal limit violations in each iteration.
}

The PDL framework for SCOPF proposed in this paper adapts the binary
search of the CCGA in its end-to-end pipeline. The adaptation is
presented in detail later in the paper.

\section{Primal-Dual Learning}
\label{sec:model}
This section reviews Primal-Dual Learning (PDL) \cite{park2023self},
the machine learning framework used to approximate the SCOPF.

\subsection{The Augmented Lagrangian Method}

Consider the following optimization problem 
\begin{equation}
\label{eq:opt}  
\begin{aligned}
\min_{\y} \obj(\y) \mbox{ subject to }  \eq(\y) = 0. 
\end{aligned}
\end{equation}
\noindent
where $\x$ represents instance parameters that determine the objective function $\obj$ and the equality constraint $\eq$. The {\em penalty method}
addresses this problem by solving a sequence of unconstrained optimization problems of the form
\begin{equation}
\label{eq:penalty}
\obj(\y) +  \rho \mathbf{1}^{\top}\viol{\eq(\y)}, 
\end{equation}
\noindent
where $\viol{\cdot}$ is the element-wise violation penalty function,
i.e., $\viol{x}= x^2$, and $\rho$ is the penalty coefficient. 
The {\em Augmented Lagrangian Method} (ALM)
\cite{powell1969method,rockafellar1974augmented,andreani2008augmented,bertsekas2014constrained}
extends the penalty method and solves unconstrained optimization
problems of the form
\begin{equation}
\label{eq:ALM}
\obj(\y) \!+\! \deq^T\eq(\y) + \frac{\rho}{2}\mathbf{1}^{\top}\viol{\eq(\y)} 
\end{equation}
\noindent
where $\deq$ are the Lagrangian multiplier approximations. 
These multipliers are updated using the rule
\begin{equation}
\label{eq:UR}
\deq \leftarrow \deq + \rho \eq(\y).
\end{equation}

\subsection{Self-Supervised Primal-Dual Learning}

\begin{figure}[!t]
\centering
\includegraphics[width=.99\columnwidth]{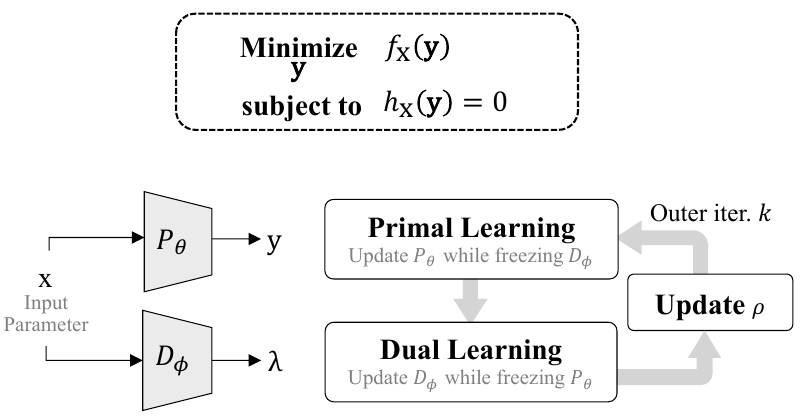}
\caption{Overview of the Self-Supervised Primal-Dual Learning.}
\label{fig:PDLoverview}
\end{figure}

PDL is a self-supervised method for training deep neural networks for
learning constrained optimization problems of the form
\eqref{eq:opt}. PDL mimics the ALM on a set of training instances,
with the aim of producing approximations to the primal and dual
optimal solutions for unseen problems from the same distribution.  An
overview of PDL is shown in Figure \ref{fig:PDLoverview}. PDL is
composed of a primal network and a dual network that are trained
iteratively in sequence to predict the primal solutions and the
Lagrangian multipliers of the ALM.

At each iteration, the {\em primal learning} step updates the
parameters $\theta$ of the primal network $\pnet$ while keeping the
dual network $\dnet$ fixed. 
Given the output of the frozen dual network $\deq$ and the penalty coefficient $\rho$, the primal learning uses the loss function
\begin{equation}\label{eq:pdl_primal_loss}
\begin{aligned}
\cL_{\text{p}}\left( \y \middle| \deq, \rho \right) \!=\! \obj(\y) \!+\! \deq^T\eq(\y) 
+ \frac{\rho}{2}\mathbf{1}^{\top}\viol{\eq(\y)}, 
\end{aligned}
\end{equation}
\noindent
which is the direct counterpart of the unconstrained optimization
\eqref{eq:ALM}.  After completion of the primal learning, PDL applies
a {\em dual learning} step that updates the parameters $\phi$ of the
dual network $\dnet$. The dual learning training uses the loss function
\begin{equation}\label{eq:pdl_dual_loss}
\begin{aligned}
    \cL_{\text{d}}\!\left(\!\deq \middle| \y, \deq_k, \rho \!\right) \!&=\!\norm{\deq \!-\! \left(\deq_k \!+\! \rho \eq(\y)\right)},
\end{aligned}
\end{equation}
which is the direct counterpart of the update rule for the Lagrangian
multipliers of the ALM~\eqref{eq:UR}.
\tb{ 
Unlike the ALM, the Lagrangian multipliers are the output returned by
the dual network. In order to update the Lagrangian multipliers, PDL
employs the \emph{copied dual network} $D_{\phi_k}$ to obtain the
Lagrangian multipliers $\deq_k$ of the current outer iteration $k$ and
update the dual network $D_{\phi_k}$ using the loss
function~\eqref{eq:pdl_dual_loss}. This approach maintains the
scalability of the learning process by avoiding to store the dual
values for all training instances.}

To handle severe violations, each iteration may increase the penalty
coefficient $\rho$. 
At iteration $k$, this update uses the maximum violation $v_k$ defined as
\begin{equation}
\label{eq:rho_sigma}
    v_k = \max_{\x\sim\dataset}\left\{\linfnorm{\eq(\y)}\right\},
\end{equation}
where $\dataset=\{\x^{(i)}\}_{i=1}^N$ is the training dataset.  The
penalty coefficient increases when the maximum violation $v_k$ is
greater than a tolerance value $\tau$ times the maximum violation from
the previous iteration $v_{k-1}$, i.e., 
\begin{equation}
\label{eq:update_rho-main}
\rho \leftarrow \min\{\alpha\rho,\rhomax\}\;\text{if } v_k > \tau v_{k-1},
\end{equation}
where $\tau\in(0,1)$ is the tolerance, $\alpha\!>\!1$ is an update
multiplier, and $\rhomax$ is an upper bound on the penalty
coefficient. 
In other words, this update process increases the
penalty coefficient only when the violation exceeds the specified
tolerance, up to the maximum value $\rhomax$.

\begin{algorithm}[!t]
\caption{Primal-Dual Learning (PDL) \cite{park2023self}}\label{alg:pdl}
\textbf{Parameter}: 
    Initial penalty coefficient $\rho$, 
    Maximum outer iteration $K$, 
    Maximum inner iteration $L$, 
    Penalty coefficient updating multiplier $\alpha$, 
    Violation tolerance $\tau$, 
    Upper penalty coefficient safeguard $\rhomax$ \\ 
\textbf{Input}: Training dataset $\dataset$\\
\textbf{Output}: learned primal and dual nets $\pnet$, $\dnet$
\begin{algorithmic}[1] 
\For{$k \in \{1,\dots,K\}$}
    \For{$l \in \{1,\dots,L\}$} \Comment{Primal Learning}
        \State Update $\pnet$ using $\nabla_\theta \cL_{\text{p}}$ (See Eq.~\eqref{eq:pdl_primal_loss})
    \EndFor
    \State Calculate $v_k$ as Eq.~\eqref{eq:rho_sigma}
    \State Define $D_{\phi_k}$ by copying $\dnet$
    \For{$l \in \{1,\dots,L\}$} \Comment{Dual Learning}
        \State Update $\dnet$ using $\nabla_\phi \cL_{\text{d}}$ (See Eq.~\eqref{eq:pdl_dual_loss})
    \EndFor
    \State Update $\rho$ using Eq.~\eqref{eq:update_rho-main}
\EndFor
\State \textbf{return} $P_{\theta}$ and $D_{\phi}$
\end{algorithmic}
\end{algorithm}

The overall PDL procedure is specified in Algorithm~\eqref{alg:pdl}, and
alternates between updating the primal and dual networks during
training. This iterative process resembles the ALM for constrained
optimization and replaces the unconstrained optimizations and the
Lagrangian updates in each outer iteration by the training of the
primal and dual networks. At each iteration, these networks
approximate the solutions of the unconstrained optimizations and the
Lagrangian multipliers for all instances in the training set. At
inference time, these networks return approximations of the primal and
dual optimal solutions for unseen problem instances.

\section{Primal-Dual Learning for SCOPF}
\label{section:PDL-SCOPF}
\begin{figure}[!t]
\centering
\includegraphics[width=.99\columnwidth]{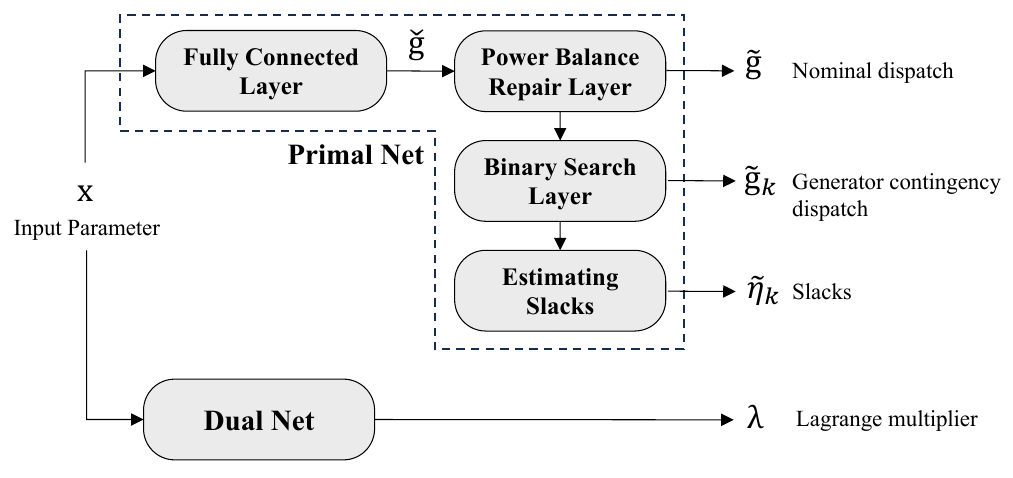}
\caption{The Primal and Dual Networks of PDL-SCOPF.}
\label{fig:schematic}
\end{figure}

This section describes PDL-SCOPF, the framework that applies PDL to
learn large-scale SCOPFs. The primal variables to approximate for the
SCOPFS are $\y:=\yapp,
\{\gkapp,\tilde{n}_k,\tilde{\rho}_k\}_{k\!\in\!\Kg},
\{\slackapp_k\}_{\{0\}\cup\Kg\cup\Ke}$, the objective function
$\obj(\y)$ is the original objective function \eqref{eq:scopf_ext_obj} of Model
\ref{model:scopf_ext}, and the constraints $\eq(\y)$ capture the power
balance equations \eqref{eq:scopf_ext_cnst_pb_kg} for the generator
contingencies as explained shortly. Figure~\ref{fig:schematic}
provides a schematic representation of the primal and dual networks
which, given the input configuration vector $\x$, estimate the primal
and dual solutions for SCOPF, respectively. There are two key
innovations is the design of the primal learning network of PDL-SCOPF:
(1) the use of a repair layer for restoring the power balance of the
base case and (2) the use of a binary search layer to estimate the
generator dispatches in the generator contingencies.

\subsection{The Primal Network}

The primal network estimates the nominal dispatch, the contigency
dispatches, and the slacks of the thermal constraints. As shown in
Figure~\ref{fig:schematic}, it uses three main components: (1) a fully
connected layer that produces a first approximation ${\bf \hat{g}}$ of
the nominal dispatch; (2) a power balance feasibility layer ${\bf
  \tilde{g}}$ that produces a second approximation of the nominal
dispatch that is guaranteed to satisfy the power balance constraints;
and (3) a binary search layer that computes an approximation ${\bf
  \tilde{g}_k}$ to the contingency dispatches by mimicking the binary
search of the CCGA. {\em These three components, and the computation
of the constraint slacks, constitutes a differentiable program for the
primal learning step that is trained end-to-end.}

\paragraph{The Fully Connected Layers}
The fully connected layer produces an approximation $\yappp$ of the
nominal dispatch that satisfies the generator bound constraints
\eqref{eq:scopf_ext_cnst_gbound}. This can be achieved by applying an
element-wise sigmoid function to the last layer of the fully connected
neural network \cite{pan2020deepopf,park2023self}.

\paragraph{Power Balance Repair Layer for Base Case}

The power balance repair layer is borrowed from \cite{chen2023end} and
guarantees that the primal network generates a nominal dispatch that
satisfies the power balance constraint
\eqref{eq:scopf_ext_cnst_pb}. It receives $\yappp$ as an input and
generates a new nominal dispatch $\yapp$ by applying a proportional
scaling of all generators as follows:
\begin{equation}
\label{eq:frlayer}
\yapp = \begin{cases}
            (1-\zeta^\uparrow)\yappp + \zeta^\uparrow\gub     \;\;\;\;\text{if }\mathbf{1}^{\top}\yappp<\mathbf{1}^{\top}\load, \\
            (1-\zeta^\downarrow)\yappp + \zeta^\downarrow\glb \;\;\;\;\text{otherwise,}
        \end{cases}
\end{equation}
where $\zeta^\uparrow$ and $\zeta^\downarrow$ are defined as,
\begin{align}
    \zeta^\uparrow   = \frac{\mathbf{1}^{\top}\load-\mathbf{1}^{\top}\yappp}{\mathbf{1}^{\top}\gub-\mathbf{1}^{\top}\yappp}, & &
    \zeta^\downarrow = \frac{\mathbf{1}^{\top}\yappp-\mathbf{1}^{\top}\load}{\mathbf{1}^{\top}\yappp-\mathbf{1}^{\top}\glb}.
\end{align}
\tb{ 
In the case where the total generation is smaller than the total loads, i.e., $\mathbf{1}^{\top}\yappp<\mathbf{1}^{\top}\load$, 
the coefficient $\zeta^\uparrow$ proportionally increases $\yappp$ so that the power balance for the base case is satisfied, i.e., $\mathbf{1}^{\top}\yappp=\mathbf{1}^{\top}\load$.
The direct output from the fully connected layer $\yappp$ is corrected to $\yapp$, that is 
\begin{equation}
\label{eq:frlayer_detail}
\begin{aligned}
\mathbf{1}^{\top}\yapp &= (1-\zeta^\uparrow)\mathbf{1}^{\top}\yappp + \mathbf{1}^{\top}\gub\\
&= \mathbf{1}^{\top}\yappp + \zeta^\uparrow(\mathbf{1}^{\top}\gub-\mathbf{1}^{\top}\yappp)\\
&=\mathbf{1}^{\top}\yappp + \mathbf{1}^{\top}\load - \mathbf{1}^{\top}\yappp\\
&=\mathbf{1}^{\top}\load.
\end{aligned}
\end{equation}
Similarly, when $\mathbf{1}^{\top}\yappp\ge\mathbf{1}^{\top}\load$, the power balance layer is designed to satisfy the power balance constraint for the base case.
For more details, please refer to~\cite{chen2023end}.
This power balance layer is differentiable almost everywhere and can thus be naturally integrated into the whole training process.
}

\paragraph{The Binary Search Layer for Generator Contingencies}

\begin{algorithm}[!t]
\caption{The Binary Search Layer \texttt{BSLayer}($\g$,$k$)}
\label{alg:bslayer}
{\scriptsize
\textbf{Parameter}:  Maximum iteration $t$\\
Initialize: $n_k\!=\!0.5$, $n_{\text{min}}\!=\!0$, $n_{\text{max}}\!=\!1$
\begin{algorithmic}[1] 
\For{$j=0,1,\dots,t$}
    \State $g^{(j)}_{k\!,i} \leftarrow \min\{g_i\!+\!n_k\gamma_i\hat{g}_i, \overline{g}_i\}$, $\forall i\in\cG$
    \State $g^{(j)}_{k\!,k} \leftarrow 0$
    \State $e_k\leftarrow \mathbf{1}^{\top}\!\mathbf{g}^{(j)}_k \!-\! \mathbf{1}^{\top}\!\load$
    \State \textbf{if} $e_k>0$ \textbf{then:} $n_{\text{max}}\leftarrow n_k$
    \State \textbf{else:} $n_{\text{min}}\leftarrow n_k$
    \State $n_k \leftarrow 0.5(n_{\text{max}}\!+\!n_{\text{min}})$
\EndFor
\For{$i\in\cG$}
\State \textbf{if} $g_i\!+\!n_k\gamma_i\hat{g}_i > \overline{g}_i$ \textbf{then:} $\rho_{k\!,i}\leftarrow 1$
\State \textbf{$\!$else:} $\rho_{k\!,i}\leftarrow 0$
\EndFor
\State\textbf{return} $\mathbf{g}^{(j)}_k,\nk,\rhok$
\end{algorithmic}
}
\end{algorithm}

Estimating the dispatches for all $N\!-\!1$ generator contingencies
presents a computational challenge, as the number of binary variables
grows quadratically with the number of generator contingencies. The
learning problem becomes impractical when dealing with industry-sized
power grids. To address this challenge, the primal network only
predicts the nominal dispatch and uses a binary search layer, inspired
by the CCGA algorithm, to compute the generator dispatches under all
contingencies. The {\em binary search layer} is an adaption of its
CCGA counterpart and is described in Algorithm~\ref{alg:bslayer}. It
estimates the dispatches under the generator contingencies ($\gk$) and
the APR-related variables ($\nk, \rhok$) from the nominal
dispatches. The algorithm performs a binary search on the global
signal $n_k$ in order to try to find contingency dispatches that
satisfy the power balance constraint. 
\tb{ 
Contrary to its CCGA counterpart, Algorithm~\ref{alg:bslayer} may not
always satisfy the power balance constraint in the contingencies 
because $\gkapp$ does not consider 
constraint~\eqref{eq:ccga_master_cnst_provisional}, which provides 
such a guarantee. In the experiment, it is observed that the 
provisional constraint~\eqref{eq:ccga_master_cnst_provisional} is 
violated in some cases in the early stage of the training, but it is satisfied when the training is complete.
}
Algorithm~\ref{alg:bslayer} is
described for a single training instance but it can be easily adapted
to simultaneously compute these solutions for all instances and all
generator contingencies in a minibatch. 
Note also that while the forward process of the binary search layer 
is conducted using Algorithm~\ref{alg:bslayer}, the backpropagation 
of it to compute the first derivatives 
of the contingency dispatches with respect to the base case 
dispatches is obtainable through the expression in terms of $\yapp$ 
and $n_k$ (see Eq.~\eqref{eq:apr}) which is efficient and differentiable almost everywhere.

\paragraph{Retrieving the Slack Estimates for Thermal Limits}

Once the generation dispatches for the base case and generator
contingencies are estimated, it is possible to calculate the slack
variables for the base case, the generator contingencies
\eqref{eq:scopf_ext_cnst_pf_kg}, and the line contingencies
\eqref{eq:scopf_ext_cnst_pf_ke}.  For the base case, the slack
variables can be estimated by
\begin{equation}
\label{eq:calculate_baseslack}
    \slackapp_0 = \max\{0, \f-\fub, \flb-\f \}.
\end{equation}
The slack variables associated with generator contingency $k\in\Kg$ are given by
\begin{equation}
\label{eq:calculate_genslack}
    \slackapp_k = \max\{0, \PTDF (\load\!-\!\B\gk)\!-\!\fub, \flb\!-\!\PTDF (\load\!-\!\B\gk)\}.
\end{equation}
Element $l$ of the slack variables associated with the line
contingency $k \in\Ke$ is given by
\begin{equation}
\label{eq:calculate_lineslack}
    \slackapp_k = \max\{0, \f\!+\!f_k\LODF_k\!-\!\fub, \flb\!-\!\f\!-\!f_k\LODF_k\},
\end{equation}
where $l\neq k$. If $l=k$, $\eta^e_{k,k}=0$.

\paragraph{The Relaxed Constraints}

The modeling of the original SCOPF problem \eqref{model:scopf_ext}
offers the advantageous property that several constraints are
implicitly satisfied.  As previously described, the power balance
constraint for the nominal case \eqref{eq:scopf_ext_cnst_pb} and the
generation dispatch bound \eqref{eq:scopf_ext_cnst_gbound} are
fulfilled through the power balance repair layer.  Additionally, by
using the binary search layer (Algorithm~\eqref{alg:bslayer}), all the
APR-related constraints for generator contingencies
(\ref{eq:scopf_ext_cnst_pb_kg},
\ref{eq:scopf_ext_cnst_gbound_kg}--\ref{eq:scopf_ext_cnst_apr3},
\ref{eq:scopf_ext_cnst_n_kg_bound},
\ref{eq:scopf_ext_cnst_rho_kg_bound}) are satisfied.  Furthermore, the
lower bounds for the slack variables
\eqref{eq:scopf_ext_cnst_slack_bound} are also satisfied through
\eqref{eq:calculate_baseslack}, \eqref{eq:calculate_genslack}, and
\eqref{eq:calculate_lineslack}.  Note that the thermal limit
constraints (\ref{eq:scopf_ext_cnst_pf},
\ref{eq:scopf_ext_cnst_pf_kg}, and \ref{eq:scopf_ext_cnst_pf_ke}) are
soft constraints. Hence, only the power balance constraints in the generator
contingencies may be violated. It is precisely those constraints that
are captured in the primal loss function of PDL-SCOPF, i.e.,
\begin{equation*}
\eq(\y)_k = \mathbf{1}^{\top}\!\mathbf{\tilde{g}}_k \!-\! \mathbf{1}^{\top}\!\load \;\;\; (k\in\Kg).
\end{equation*}

\subsection{The Dual Network}  

Since all the other constraints are satisfied, the dual network
produces optimal dual estimates $\deq=\dnet(\x)$ for the generator
contingency power balance constraints. Experiments were run to
evaluate whether adding the output of the primal network to the input
of the dual network would yield performance improvements. The results
were inconclusive, so the PDL-SCOPF follows the schema from Section
\ref{sec:model}.

\section{Experiments}
\label{sec:exp}
\subsection{The Experimental Settings}
\label{ssec:exp_setting}

\paragraph{Test Cases}
The effectiveness of PDL-SCOPF is assessed on six
specific cases from the Power Grid Library (PGLIB)
\cite{babaeinejadsarookolaee2019power} given in
Table~\ref{tab:case_spec}.  Note that the sizes of $\Kg$ and $\Ke$ may
differ from the actual number of generators and transmission
lines. Indeed, the experiments exclude contingency scenarios where the
generator capacity is zero or the lower limit $\underline{g}$ is
negative, which indicate the possible presence of dispatchable loads.
Regarding line contingencies, the experiments exclude transmission
lines that disconnect the network completely. These lines can be identified
using the LODF matrix $\LODF$ \cite{guo2009direct}.

\begin{table}[!t]
\centering
\small
\setlength{\tabcolsep}{4pt}
\begin{tabular}{@{}lccccccccc@{}}
\toprule
Test Case  & $|\cN|$ & $|\cG|$ & $|\cL|$ & $|\cE|$ & $\lvert\Kg\rvert$ & $\lvert\Ke\rvert$ &$\dim\!\left(\x\right)$\\
\midrule
\texttt{300\_ieee}     & 300    & 69   & 201  & 411     & 57    & 322    & 339  \\
\texttt{1354\_peg}     & 1354   & 260  & 673  & 1991    & 193   & 1430   & 1193 \\
\texttt{1888\_rte}     & 1888   & 290  & 1000 & 2531    & 290   & 1567   & 1580 \\
\texttt{3022\_goc}     & 3022   & 327  & 1574 & 4135    & 327   & 3180   & 2228 \\
\texttt{4917\_goc}     & 4917   & 567  & 2619 & 6726    & 567   & 5066   & 3753 \\
\texttt{6515\_rte}     & 6515   & 684  & 3673 & 9037    & 657   & 6474   & 5041 \\
\bottomrule
\end{tabular}
\caption{Specifications of the SCOPF Test Cases.}
\label{tab:case_spec}
\end{table}

\paragraph{Data Perturbation}

The data set includes instances obtained by perturbing the load
demands, the cost coefficients, and the upper bounds of the generation
dispatch, i.e., $x:=\{\load,\objc,\gub\}$, in the PGLIB configuration.
This generalizes earlier settings
\cite{fioretto2020predicting,donti2021dc3,park2023self}, where the
only load demand $\load$ is perturbed.  As in
\cite{fioretto2020predicting,donti2021dc3,park2023self}, the load
demands were sampled from a truncated multivariate Gaussian
distribution defined as
\begin{equation}\label{eq:perturb_load}
    \load \sim \mathcal{T}\mathcal{N}\left( \load_0,\Sigma, (1-\mu)\load_0, (1+\mu)\mu \load_0 \right),
\end{equation}
where $\load_0$ is the base load demands defined in the PGLib and $\Sigma$ is
the covariance matrix. Element $\Sigma_{ij}$ of $\Sigma$ is defined as 
$\Sigma_{ij} = \alpha \sigma_i \sigma_j$ where $\sigma_i$ and $\sigma_j$ are
proportional to $d_i$ and $d_j$ (i.e., scaled by the Z-score
representing the 95th percentile of a normal Gaussian with an unit
standard deviation), and the correlation coefficient $\alpha$ is $1$
when $i=j$ and $0.5$ otherwise. $\mu$ is set to $0.5$, meaning that
the load demands were set to be perturbed by $\pm50\%$ at most.  
To perturb the cost coefficients and the dispatch upper bounds, the
experiments use the base values already provided in PGLIB.  These base
values are then adjusted by multiplying by factors specific to each
instance.  Specifically, per instance, two independent factors were
sampled from the multivariate Gaussian distributions
$\mathcal{N}(\mathbf{1}, \Sigma)$, where $\Sigma$ uses a correlation
coefficient $\alpha$ of $0.8$.  
The cost coefficients $\objc$ are safeguarded to be nonnegative. 
The dispatch upper bounds are truncated to be greater than its lower bounds as
$\gub\leftarrow\max\{\gub,\glb+0.01\gcapa\}$.
The dimension of the input parameter $\dim(\x)$ is the same as
$2\lvert\cG\rvert+\lvert\cL\rvert$.  

\paragraph{Data Generation}
To evaluate PDF-SCOPF, ground truths of 1,000 instances are sampled from the specified
distributions.
These instances were solved using the CCGA using the
parameter settings suggested in \cite{velloso2021exact}. The penalty
coefficient for slacks $\slackpen$ is set to $1500$ as in
\cite{chen2023end}.
The code is implemented based on Pyomo \cite{hart2017pyomo}, and solved 
using Gurobi 10.0 \cite{gurobi} on 24 physical cores of a machine 
with Intel Xeon 2.7GHz with 256GB RAM.  
Table~\ref{tab:numvar} highlights the size of
the extensive SCOPF formulation~\eqref{model:scopf_ext}: it gives the
numbers of variables and constraints before and after applying the
Presolve (with the default setting) in Gurobi. Presolve cannot even
complete due to out of memory for the larger test cases.
For the \texttt{3022\_goc}, there are about 14.6 million
continuous variables and 106,600 binary variables. In the largest case
\texttt{6515\_rte}, there are about 64.9 million continuous
variables, 449,400 binary variables, and 130.7 million
constraints.   Table~\ref{tab:CCGA_time} presents the computing
times and the number of iterations of the CCGA. Observe the challenging
\texttt{6515\_rte} case, which requires at least 4648 seconds to solve.

\paragraph{Baselines}

PDL-SCOPF is evaluated against three baselines to assess its
performance. 
{\em It is important to stress that all baselines use the
  same primal neural architecture, including the repair layer for the
  power balance in the base case.}  
Only PDL-SCOPF also uses a dual network. 
The first baseline, denoted as \emph{Penalty}, is a self-supervised
framework that uses the penalty function (Eq.~\ref{eq:penalty}) as a
loss function.  The other two baselines are supervised learning (SL) frameworks. The
first supervised learning framework, \emph{Na\"ive}, uses a loss function that
minimizes the distance between the base case generation dispatch
estimates and the ground truth, defined as $\norm{\yapp -
  \mathbf{g}^*}_2$. The second supervised learning framework is inspired by the
\emph{Lagrangian Duality} (LD) approach from
\cite{fioretto2020predicting}: it combines the Na\"ive loss function
with a penalty for constraint violations, i.e., loss function
\begin{equation}
    \norm{\yapp-\mathbf{g}^*}_2+\rho\mathbf{1}^{\top}\viol{\eq(\y)}.
\end{equation}
Contrary to \cite{fioretto2020predicting}, the penalty coefficient
$\rho$ is not updated and is set to $1\mathrm{e}{3}$, which was chosen
after hyperparameter tuning.  Note that to test supervised frameworks,
a significant number of training instances need to be accumulated.
\tb{ 
For the supervised learning baselines, 10,000 training instances for the six test cases were accumulated. 
}
The results will show that solving these SCOPF problems is
time-consuming, making it prohibitive for large-scale cases. However,
it was felt important to compare the supervised and self-supervised
frameworks and demonstrate that PDL-SCOPF can achieve a comparable or
better performance without the need to solve training instances.

\begin{table}[!t]
\centering
\small
\setlength{\tabcolsep}{3pt}
\begin{tabular}{@{}lcccccc@{}}
\toprule
& \multicolumn{3}{c}{Before} & \multicolumn{3}{c}{After}\\
Test Case  & \#CV & \#BV & \#Cnst & \#CV & \#BV & \#Cnst \\
\midrule
\texttt{300\_ieee} & 160.2k & 3.2k   & 327.9k & 44.5k    & 3.2k     & 82.0k \\
\texttt{1354\_peg} & 3.3m   & 50.0k  & 6.7m   & 642.3k   & 50.0k    & 642.3k\\
\texttt{1888\_rte} & 4.8m   & 83.8k  & 9.7m   & 503.4k   & 83.8k    & 649.7k\\
\texttt{3022\_goc} & 14.6m  & 106.6k & 29.4m  & -         & -         & - \\
\texttt{4917\_goc} & 38.2m  & 321.5k & 77.1m  & -         & -         & - \\
\texttt{6515\_rte} & 64.9m  & 449.4k & 130.7m & -         & -         & - \\
\bottomrule
\end{tabular}
\caption{The Number of Binary and Continuous Variables Denoted as \#BV
  and \#CV and Constraints (\#Cnst) in Extensive SCOPF Problem Before
  and After Presolve. `$-$' indicates that Presolve runs of out of
  memory. $k$ and $m$ signify $10^3$ and $10^6$.}
\label{tab:numvar}
\end{table}

\begin{table}[!t]
\centering
\small
\setlength{\tabcolsep}{4pt}
\begin{tabular}{@{}lcccccc@{}}
\toprule
           & \multicolumn{3}{c}{Solving Time (s)} & \multicolumn{3}{c}{\#Iteration} \\
              \cmidrule(lr){2-4}\cmidrule(lr){5-7}
Test Case  & min & mean & max         & min & mean & max \\
\midrule
\texttt{300\_ieee}     & 7.72    & 32.06 & 99.10 & 4 & 5.06 & 8 \\
\texttt{1354\_peg}     & 42.66   & 97.10 & 1125.91 & 3 & 4.79 & 7 \\
\texttt{1888\_rte}     & 171.92  & 654.39 & 2565.6 & 3 & 4.93 & 6 \\
\texttt{3022\_goc}     & 628.67  & 6932.20 & 15844.52 & 8 & 10.71 & 17 \\
\texttt{4917\_goc}     & 3720.33 & 8035.86 & 15316.95 & 7 & 10.98 & 16 \\
\texttt{6515\_rte}     & 4648.31 & 9560.78 & 92430.47 & 4 & 6.65 & 10 \\
\bottomrule
\end{tabular}
\caption{Elapsed Time and Iterations for Solving SCOPF Test Instances Using CCGA.}
\label{tab:CCGA_time}
\end{table}

\paragraph{Architectural Details}

Both the primal and dual networks consist of four fully-connected
layers, each followed by Rectified Linear Unit (ReLU) activations.
Layer normalization \cite{ba2016layer} is applied before the
fully-connected layers for the primal network only.  The number of
hidden nodes in each fully-connected layer is proportional to the
dimension of the input parameter, and is set to be $1.5\dim(\x)$.

\paragraph{Training Setting}

The training uses a mini-batch size of 8 and a maximum of 1,000
epochs.  The models are trained using the Adam optimizer
\cite{kingma2014adam} with a learning rate of
$1\mathrm{e}{\textrm{-}4}$, which is reduced by a factor of 0.1 at
90\% of the total iterations.  For PDL-SCOPF, the number of outer
iterations ($K$) is set to 20, and the maximum number of inner
iterations ($L$) is set to 2,000 for both the primal and dual
networks, resulting in a total of $80,000$ iterations.  This same
iteration count is applied to the other baselines.  The implementation
uses PyTorch, and all models were trained on a machine equipped with a
NVIDIA Tesla V100 GPU and an Intel Xeon 2.7GHz CPU. Averaged
performance results based on five independent training processes
with different seeds are reported.

\paragraph{Parameter Settings}
The PDL-SCOPF parameters are configured following the recommendations
in \cite{park2023self}.  A minor adjustment relates to updating the
dual net.  When the penalty coefficient becomes too high, updating the
dual net can be unstable due to the substantial gap between the
desired dual estimates and the current estimates in
Eq.~\eqref{eq:pdl_dual_loss}.  To address this, the penalty
coefficient used in the loss function (Eq.~\ref{eq:pdl_dual_loss}) is
fixed at $1\mathrm{e}{\textrm{-}1},$ regardless of how the penalty
coefficient is updated (Eq.~\ref{eq:update_rho-main}).  The initial
penalty coefficient used in the primal loss function
(Eq.~\ref{eq:pdl_primal_loss}) is set to $0.1$ and may be increased up
to $\rhomax=1\mathrm{e}{8}$.  For updating the penalty coefficient,
$\tau$ and $\alpha$ are set to $0.9$ and $2.0$, respectively. To
balance the objective function value and the constraint violations in
the primal loss function (Eq.~\ref{eq:pdl_primal_loss}), the objective
function is divided by $1\mathrm{e}{5}$.

\subsection{Numerical Results}

\begin{table}[!t]
\centering
\small
\begin{tabular}{@{}l|cccc@{}}
\toprule
\begin{tabular}{@{}c@{}}\\ Test Case\end{tabular}
& \begin{tabular}{@{}c@{}}Na\"ive \\ (SL)\end{tabular} 
& \begin{tabular}{@{}c@{}}LD \\ (SL)\end{tabular} 
& \begin{tabular}{@{}c@{}} Penalty \\ (SSL)\end{tabular} 
& \begin{tabular}{@{}c@{}}PDL-SCOPF \\ (SSL) \end{tabular}\\
\midrule
\texttt{300\_ieee} & 0.008& 0.001 & 0.003 & 0.000 \\
\texttt{1354\_peg} & 0.022& 0.003 & 0.000 & 0.000 \\
\texttt{1888\_rte} & 0.044& 0.003 & 0.000 & 0.000 \\
\texttt{3022\_goc} & 0.088& 0.006 & 0.000 & 0.000 \\
\texttt{4917\_goc} & 0.067& 0.002 & 0.000 & 0.000 \\
\texttt{6515\_rte} & 0.001& 0.000 & 0.000 & 0.000 \\
\bottomrule       
\end{tabular}
\caption{Maximum Violations on Power Balance Constraints for Generator Contingency (in p.u.).}
\label{tab:pb_kg}
\end{table}

\begin{table}[!t]
\centering
\small
\setlength{\tabcolsep}{3pt}
\begin{tabular}{@{}l|r@{}lr@{}lr@{}lr@{}l@{}}
\toprule
\begin{tabular}{@{}c@{}}\\ Test Case\end{tabular}
& \multicolumn{2}{c}{ \begin{tabular}{@{}c@{}} Na\"ive \\ (SL)\end{tabular} } 
& \multicolumn{2}{c}{ \begin{tabular}{@{}c@{}} LD \\ (SL)\end{tabular} }
& \multicolumn{2}{c}{ \begin{tabular}{@{}c@{}} Penalty \\ (SSL)\end{tabular} } 
& \multicolumn{2}{c}{ \begin{tabular}{@{}c@{}} PDL-SCOPF \\ (SSL) \end{tabular} }\\
\midrule
\texttt{300\_ieee} & 5&(8.77\%)  & 2&(3.51\%)  & 2&(3.51\%) & 0&(0.00\%) \\
\texttt{1354\_peg} & 14&(7.25\%) & 4&(2.07\%)  & 0&(0.00\%) & 0&(0.00\%) \\
\texttt{1888\_rte} & 31&(10.68\%)& 7&(2.41\%)  & 0&(0.00\%) & 0&(0.00\%) \\
\texttt{3022\_goc} & 57&(17.43\%)& 13&(3.98\%) & 0&(0.00\%) & 0&(0.00\%) \\
\texttt{4917\_goc} & 42&(7.41\%) & 12&(2.12\%) & 0&(0.00\%) & 0&(0.00\%) \\
\texttt{6515\_rte} & 2&(0.30\%)  & 0&(0.00\%)  & 0&(0.00\%) & 0&(0.00\%) \\
\bottomrule       
\end{tabular}
\caption{The Number of Generator Contingencies with Violated Power Balance Constraints (Percentages in Parenthesis).
}
\label{tab:pb_kg_freq}
\end{table}

The numerical results compare PDL-SCOPF with the ground truth and the
baselines.
\tb{ 
  First, to check whether the
  constraints are satisfied by the optimal solution estimates from
  ML-based methods, Tables~\ref{tab:pb_kg}--\ref{tab:pb_kg_freq} show
  the violations of the hard constraints, that is, the power balance
  constraint for generator
  contingencies~\eqref{eq:scopf_ext_cnst_pb_kg}.
  Table~\ref{tab:pb_kg} reports the averaged maximum violation of the
  generator contingency power balance constraints on testing
  instances.  PDL-SCOPF and the SSL penalty method have negligible
  violations of the power balance constraints on all instances.  In
  contrast, the supervised methods exhibit violations that can be
  significant, especially for the Na\"ive SL baseline.}
\tb{  Table~\ref{tab:pb_kg_freq} shows the corresponding
number of generator contingencies where the power balance constraint
is violated and its percentages over the whole number of generator
contingencies in parenthesis. The violation are calculated with a
tolerance of $1\mathrm{e}{\textrm{-}4}$.  This table also underscores
PDL-SCOPF satisfies the hard constraints, which was a primary goal of
the proposed training procedure.}

\begin{figure}[!t]
\centering
\includegraphics[width=.99\columnwidth]{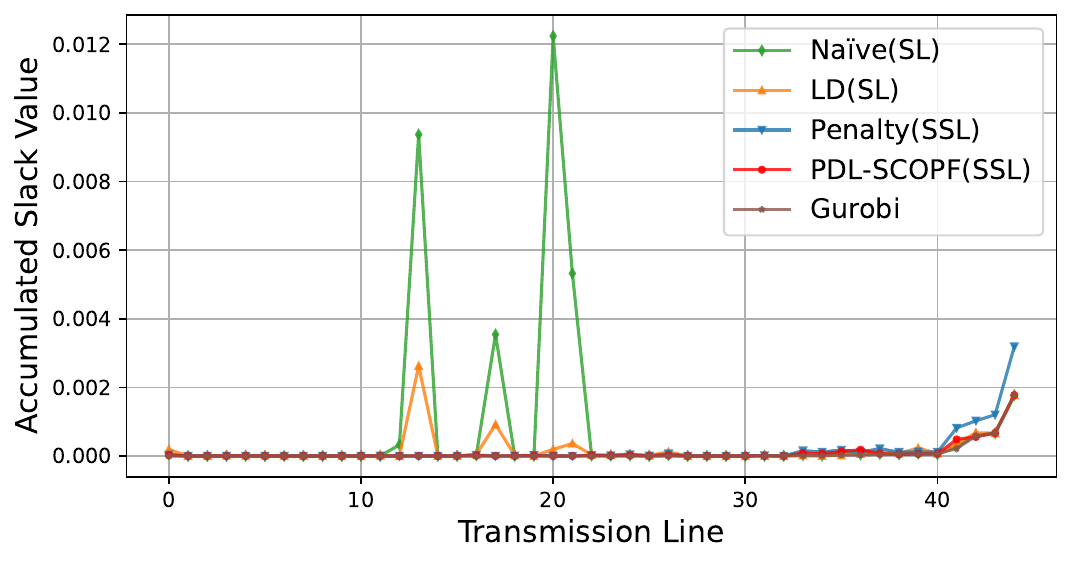}
\caption{\tb{Averaged Total Power Balance Constraint Slack Values for Each Transmission Line in \texttt{1354\_pegase}.}}
\label{fig:slack}
\end{figure}

\tb{ 
Second, Table~\ref{tab:optgap} reports the mean optimality gap in 
percentage, providing a comparison between the CCGA algorithm and 
its learning counterparts. The results show that PDL-SCOPF is 
almost always the strongest method with optimality gaps often 
below 1\%. It is only dominated on the IEEE 300-bus system by LD 
but this result must be treated carefully since LD violates some 
of the power balance constraints.  Note that PDL-SCOPF 
significantly dominates the Penalty (SSL) method. Also, for the 
thermal limits for the base case and contingencies, it is 
desirable to have smaller slack variable values. For instance of 
\texttt{1354\_pegase}, Figure~\ref{fig:slack} represents the 
the sum of the slack values for the base case, the 193 generation 
contingencies, and the 1430 transmission contingencies. It only shows the 
slack variable values for the transmission lines in which 
only positive values are observed. The transmission lines are 
sorted by the total slack values for Na\"ive SL baseline in 
ascending order. 
PDL-SCOPF exhibits the smallest slack values when compared with the 
baseline and ground truth (Gurobi).
Note that positive slack values are quite sparsely observed 
only on a few transmission lines for PDL-SCOPF, meaning that the 
transmission thermal limits are satisfied on most transmission 
lines for the contingencies. 
This result also contributes to the small optimality gaps of 
PDL-SCOPF (Table~\ref{tab:optgap}) because  
non-zero slack values lead to suboptimal solutions given their high penalties in the objective. 
}

\begin{table}[!t]
\centering
\small
\begin{tabular}{l|cccc}
\toprule
\begin{tabular}{@{}c@{}}\\ Test Case\end{tabular}
& \begin{tabular}{@{}c@{}}Na\"ive\\ (SL)\end{tabular} 
& \begin{tabular}{@{}c@{}}LD\\ (SL)\end{tabular} 
& \begin{tabular}{@{}c@{}} Penalty \\ (SSL)\end{tabular} 
& \begin{tabular}{@{}c@{}}PDL-SCOPF \\ (SSL) \end{tabular}\\
\midrule
\texttt{300\_ieee}     & 1.021   & \textbf{0.908} & 3.867 & 2.805 \\
\texttt{1354\_peg}     & 13.447  & 2.700 & 2.533 & \textbf{0.856} \\
\texttt{1888\_rte}     & 22.115  & 2.436 & 4.969 & \textbf{1.960} \\
\texttt{3022\_goc}     & 159.116 & 8.008 & 1.312 & \textbf{0.983} \\ 
\texttt{4917\_goc}     & 47.212  & 2.096 & 0.454 & \textbf{0.210} \\
\texttt{6515\_rte}     & 2.419   & 1.292 & 2.069 & \textbf{0.815} \\
\bottomrule       
\end{tabular}
\caption{Mean Optimality Gap (\%) (best values in bold).}
\label{tab:optgap}
\end{table}

\begin{table}[!t]
\centering
\small
\begin{tabular}{l|rr}
\toprule
Test Case & Training & Sampling\\
\midrule
\texttt{300\_ieee} & 26min & 89hr \phantom{0}3min\\
\texttt{1354\_peg} & 36min & 269hr 43min\\
\texttt{1888\_rte} & 43min & 1817hr 45min\\
\texttt{3022\_goc} & 1hr \phantom{0}7min & 19256hr \phantom{0}7min\\
\texttt{4917\_goc} & 1hr 59min & 22321hr 50min\\
\texttt{6515\_rte} & 2hr 54min & 26557hr 43min\\
\bottomrule       
\end{tabular}
\caption{\tb{Averaged Training Time (in GPU) for PDL-SCOPF and Accumulated CPU Time to Prepare the Supervised Training Dataset.}}
\label{tab:time}
\end{table}

\begin{table}[!t]
\centering
\small
\setlength{\tabcolsep}{3pt}
\begin{tabular}{l|cccc|c}
\toprule
            & \multicolumn{4}{c|}{PDL-SCOPF Inference Time (ms)} & \multirow{2}{*}{Speedup}\\
Test Case  & 8@GPU   &  1@GPU   &  8@CPU     & 1@CPU   & \\
\midrule                                                         
\texttt{300\_ieee} & 5.163   &  5.117   &  12.392    & 3.527   & 9088.23$\times$ \\
\texttt{1354\_peg} & 6.917   &  5.146   &  290.376   & 25.745  & 3771.60$\times$ \\ 
\texttt{1888\_rte} & 7.956   &  5.145   &  455.846   & 46.876  & 13960.02$\times$ \\
\texttt{3022\_goc} & 13.486  &  5.823   &  1484.019  & 183.879 & 37699.81$\times$\\
\texttt{4917\_goc} & 31.775  &  8.043   &  3975.468  & 479.881 & 16745.53$\times$\\
\texttt{6515\_rte} & 51.9370 &  10.576  &  6767.474  & 823.016 & 11616.76$\times$\\
\bottomrule       
\end{tabular}
\caption{
\tb{ 
Averaged Inference Time of PDL-SCOPF with 1 Instance or 8 Instances on CPU or GPU in Milliseconds and Averaged Speedup over Gurobi on 1 CPU.
}
}
\label{tab:inference_time}
\end{table}

\tb{ 
Table~\ref{tab:time} reports the elapsed GPU times for PDL-SCOPF training and the 
accumulated CPU time for generating samples for the supervised baselines. 
The time to generate solutions offline is extremely high, even with the CCGA. 
The self-supervised learning including PDL-SCOPF does not necessitate the ground truth sampling process, thus brings substantial benefits in training time.
}
It is remarkable that, for the largest test case \texttt{6515\_rte}, the training 
time of 2 hours and 54 minutes is significantly smaller than the time required to 
solve the single worst-case SCOPF instance. The latter takes approximately 25 hours
and 40 minutes (92430.47 seconds equivalently, as shown in 
Table~\ref{tab:CCGA_time}) in the test dataset. This efficiency in training 
can be attributed to the scalable modeling and 
the self-supervised nature of PDL-SCOPF.

Table~\ref{tab:inference_time} reports the inference time in
milliseconds of PDL-SCOPF on CPUs and GPUs for a single instance or a
batch of 8 instances. Even for the largest test case, PDL-SCOPF
provides a high-quality approximation to a single instance in about 10
milliseconds and to a batch of instances in about 50
milliseconds. Note also the significant benefits of using
GPUs. \tb{These results show that PDL-SCOPF is 4 orders of magnitude
  faster than Gurobi on these instances.}

{\em Altogether, these results show that PDL-SCOPF, once trained,
  provides near-optimal solutions to SCOPF in milliseconds without the
  need to solve any instance offline. It provides operators with a new tool to inform real-time decision making.}

\section{Conclusion}
\label{sec:conclusion}
This paper introduces a self-supervised primal-dual learning framework
PDL-SCOPF to approximate the optimal solutions for large-scale SCOPF
problems.  The framework produces near-optimal feasible solutions in
just milliseconds, and is self-supervised, precluding the need to
generate optimal solutions to the training instances.  PDL-SCOPF is a
primal-dual learning method that mimics the ALM framework, alternating
between approximating unconstrained optimization problems using a
primal network and estimating the Lagrangian multipliers with a dual
networks. The primal network is trained with a loss function that
combines the original objective of the SCOPF with constraints
capturing the power balance of the generator contingencies. It
features a repair layer to ensure the feasibility of the power balance
of the nominal case, and a binary search layer (inspired by the CCGA)
that computes the contingency dispatches from the nominal
dispatch. The whole framework is trained end to end, in the ALM style.

Experiment results conducted on a range of test cases, featuring
systems with up to approximately 6,500 generators, demonstrated the
effectiveness of the proposed methodology.  By combining
self-supervised learning, primal dual learning, and implicit
feasibility and completion layers, PDL-SCOPF produces predictions
with no constraints violations and small optimality gaps that compare
dominates existing supervised and self-supervised methods. 

This research represents an important step in demonstrating the
scalability of self-supervised end-to-end optimization learning
frameworks. These frameworks have the potential to transform how to
learn large-scale optimization problems.  In the context of SCOPF
problems, potential applications include corrective
SCOPF \cite{wang2016solving}, stochastic
SCOPF \cite{sharifzadeh2016stochastic}, and chance constraint SCOPF
problems \cite{roald2015security}. Self-supervised end-to-end learning
may also be an avenue for large-scale optimization challenges in
expansion planning and topology optimization.
\section*{Acknowledgments}

This research is partly supported by NSF under Awards 2007095
and 2112533.

\bibliographystyle{IEEEtran} 
\bibliography{refs}

\end{document}